\documentclass[letterpaper, 10 pt, conference]{ieeeconf}  % Comment this line out if you need a4paper

\IEEEoverridecommandlockouts                              % This command is only needed if 
\title{\LARGE \bf
Multimodal Object Query Initialization for 3D Object Detection
}

\author{Mathijs R. van Geerenstein$^{1,2,*}$,
Felicia Ruppel$^{1,3,*}$,
Klaus Dietmayer$^{3}$
and Dariu M. Gavrila$^{2}$
\thanks{*Equal contribution}
\thanks{$^{1}$Robert Bosch GmbH, Corporate Research, 71272 Renningen, Germany, 
	{\tt\footnotesize \{firstname.lastname\}@de.bosch.com}}%
\thanks{$^{2}$Intelligent Vehicles Group, Delft University of Technology, the Netherlands, 
	{\tt\footnotesize \{initials.lastname\}@\{student.\}tudelft.nl}}%
\thanks{$^{3}$Institute of Measurement, Control and Microtechnology, Ulm University, Germany,
	{\tt\footnotesize \{firstname.lastname\}@uni-ulm.de}}%
}

\usepackage{graphicx}
\usepackage{amsmath}
\usepackage{amssymb}
\usepackage{booktabs}
\usepackage{hyperref}
\usepackage[capitalize]{cleveref}
\crefname{section}{Sec.}{Secs.}
\Crefname{section}{Section}{Sections}
\Crefname{table}{Table}{Tables}
\crefname{table}{Tab.}{Tabs.}

\usepackage[table]{xcolor}
\usepackage{xcolor}
\usepackage{soul}
\usepackage{arydshln}
\usepackage{adjustbox}
\definecolor{gray1}{rgb}{0.95,0.95,0.95}
\definecolor{gray2}{rgb}{0.9,0.9,0.9}
\definecolor{light-gray}{gray}{0.55}
\definecolor{L}{rgb}{0.50980,0.70196,0.4}
\definecolor{C}{rgb}{0.588235,0.450980,0.6509803}

\usepackage{cite}

\definecolor{violet}{rgb}{0.5, 0.0, 0.5}%

\usepackage{fancyhdr}
\usepackage{ragged2e}

\fancypagestyle{FirstPage}{
	
	\lfoot{\noindent\fontsize{7.5}{8.5}\selectfont\justifying This work has been submitted to the IEEE for possible publication. Copyright may be transferred without notice, after which this version may no longer be accessible.}%\textcopyright\ 2024 IEEE.  Personal use of this material is permitted.  Permission from IEEE must be obtained for all other uses, in any current or future media, including reprinting/republishing this material for advertising or promotional purposes, creating new collective works, for resale or redistribution to servers or lists, or reuse of any copyrighted component of this work in other works.}
	
}
\makeatletter
\let\NAT@parse\undefined
\makeatother

\begin{document}

\maketitle
\thispagestyle{empty}
\pagestyle{empty}

%%%%%%%%%%%%%%%%%%%%%%%%%%%%%%%%%%%%%%%%%%%%%%%%%%%%%%%%%%%%%%%%%%%%%%%%%%%%%%%%
\begin{abstract}

3D object detection models that exploit both LiDAR and camera sensor features are top performers in large-scale autonomous driving benchmarks. 
A transformer is a popular network architecture used for this task, in which so-called object queries act as candidate objects. 
Initializing these object queries based on current sensor inputs is a common practice. 
For this, existing methods strongly rely on LiDAR data however, and do not fully exploit image features.
Besides, they introduce significant latency.
To overcome these limitations we propose EfficientQ3M, an efficient, modular, and multimodal solution for object query initialization for transformer-based 3D object detection models.
The proposed initialization method is combined with a ``modality-balanced'' transformer decoder where the queries can access all sensor modalities throughout the decoder. 
In experiments, we outperform the state of the art in transformer-based LiDAR object detection on the competitive nuScenes benchmark and showcase the benefits of input-dependent multimodal query initialization, while being more efficient than the available alternatives for LiDAR-camera initialization.
The proposed method can be
applied with any combination of sensor modalities as input,
demonstrating its modularity.
% DG removed due to length abstract
%The proposed initialization \DG{is modular and} can be applied to any combination of sensor modalities as input.

\end{abstract}

%%%%%%%%%%%%%%%%%%%%%%%%%%%%%%%%%%%%%%%%%%%%%%%%%%%%%%%%%%%%%%%%%%%%%%%%%%%%%%%%
\section{INTRODUCTION}
\thispagestyle{FirstPage}
3D object detection is a vital part of autonomous driving systems, and the resulting detections serve as a starting point for downstream tasks such as tracking or trajectory prediction. 
In automotive scenes, we try to predict multi-class 3D bounding boxes for all road users and other important objects around the ego vehicle.
Models that exploit multimodal sensor data are currently top performers in popular benchmarks for 3D object detection~\cite{caesar_nuscenes_2020, sun_scalability_2020}.
The sensor suite typically consists of a roof-mounted LiDAR and a set of monocular cameras, the former providing a sparse 3D point cloud and the latter high-resolution dense images. 
These two sensor types are complementary: LiDAR brings accurate depth information and the cameras offer texture information and higher resolution for small, far-away objects.

\begin{figure}[t]
    \centering
    \includegraphics[height=6cm]{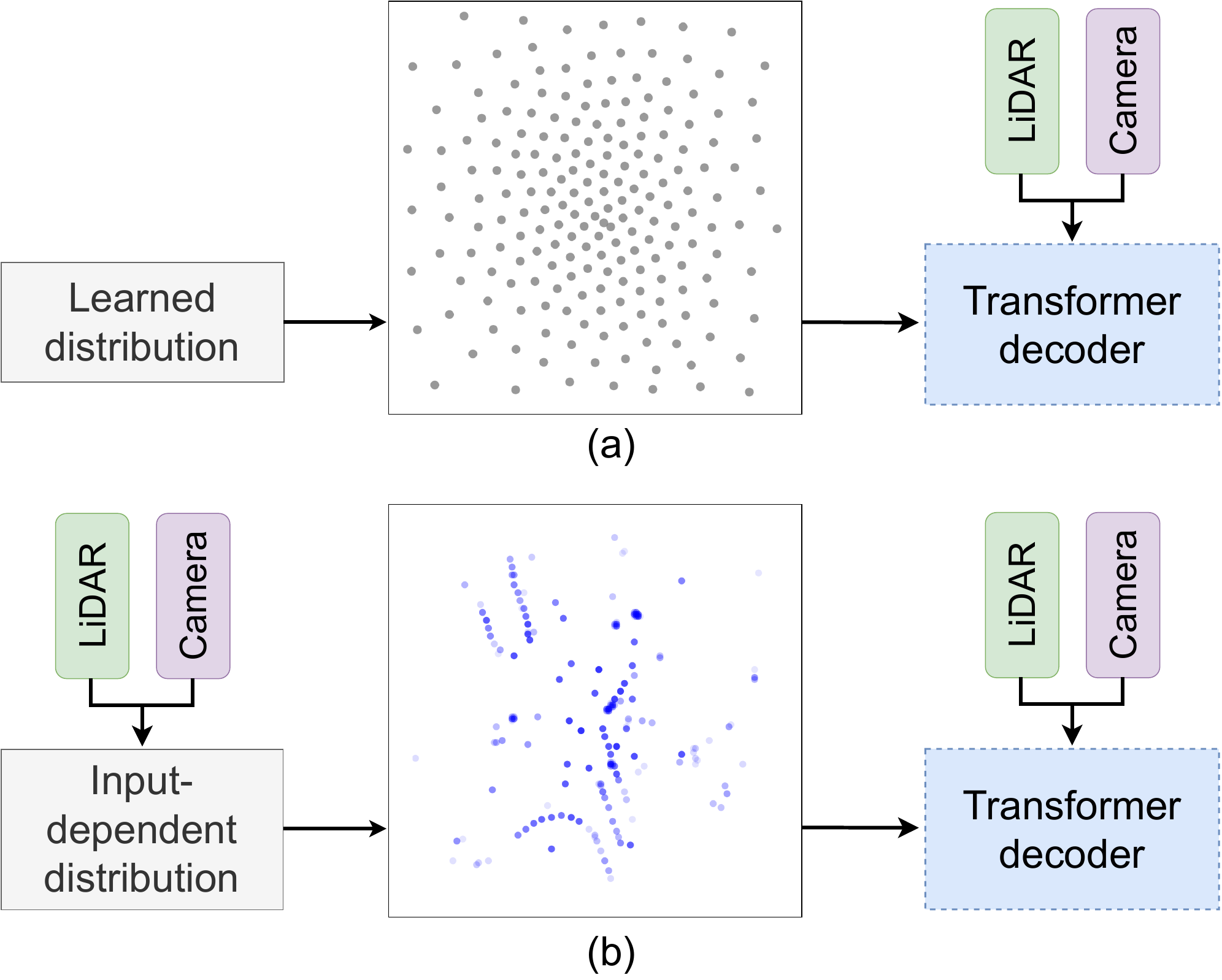}
    \caption{Different query initialization approaches in transformer-based LiDAR-camera object detection. We show the initial object query locations from the bird's eye view with
    (a) input-agnostic initialization as in FUTR3D~\cite{chen_futr3d_2023} (b) the proposed feature-informed initialization.
    % \textbf{\color{L}L} is LiDAR and \textbf{\color{C}C} is camera.
    }
    \label{fig:model_compare}
\end{figure}

In recent years, the transformer~\cite{vaswani_attention_2017} architecture has successfully been applied to 2D~\cite{carion_end--end_2020, zhu_deformable_2021, yao_efficient_2021} and 3D~\cite{misra_end--end_2021, ruppel_transformers_2022-1, bai_transfusion_2022, chen_futr3d_2023} object detection tasks.
% Besides performance improvements, their architecture provides a natural extension to the object tracking task~\cite{meinhardt_trackformer_2022, ruppel_transformers_2022}, and enables end-to-end training because it does not require non-maximum suppression (NMS) post-processing to remove duplicate predictions~\cite{carion_end--end_2020}.
Transformers rely on \emph{object queries} to detect objects, where each query is an object candidate that can detect at most one object. 
A query is essentially a feature vector in a latent space that encodes all information needed to predict a classified bounding box. 
Usually, each query is accompanied by a reference or anchor point, with respect to which the bounding box is predicted~\cite{mao_3d_2023}. 
The initialization of the query feature vectors and their locations is an active research topic, and we distinguish two approaches: learning a fixed, input-agnostic distribution for the object queries during training, or initializing them based on the current sensor inputs. 
\cref{fig:model_compare} shows the learned distribution of initial query locations from input-agnostic method FUTR3D~\cite{chen_futr3d_2023} (a), and input-dependent initialization using our proposed method (b).

\begin{figure}[htb]
    \centering
    \includegraphics[width=0.9\columnwidth]{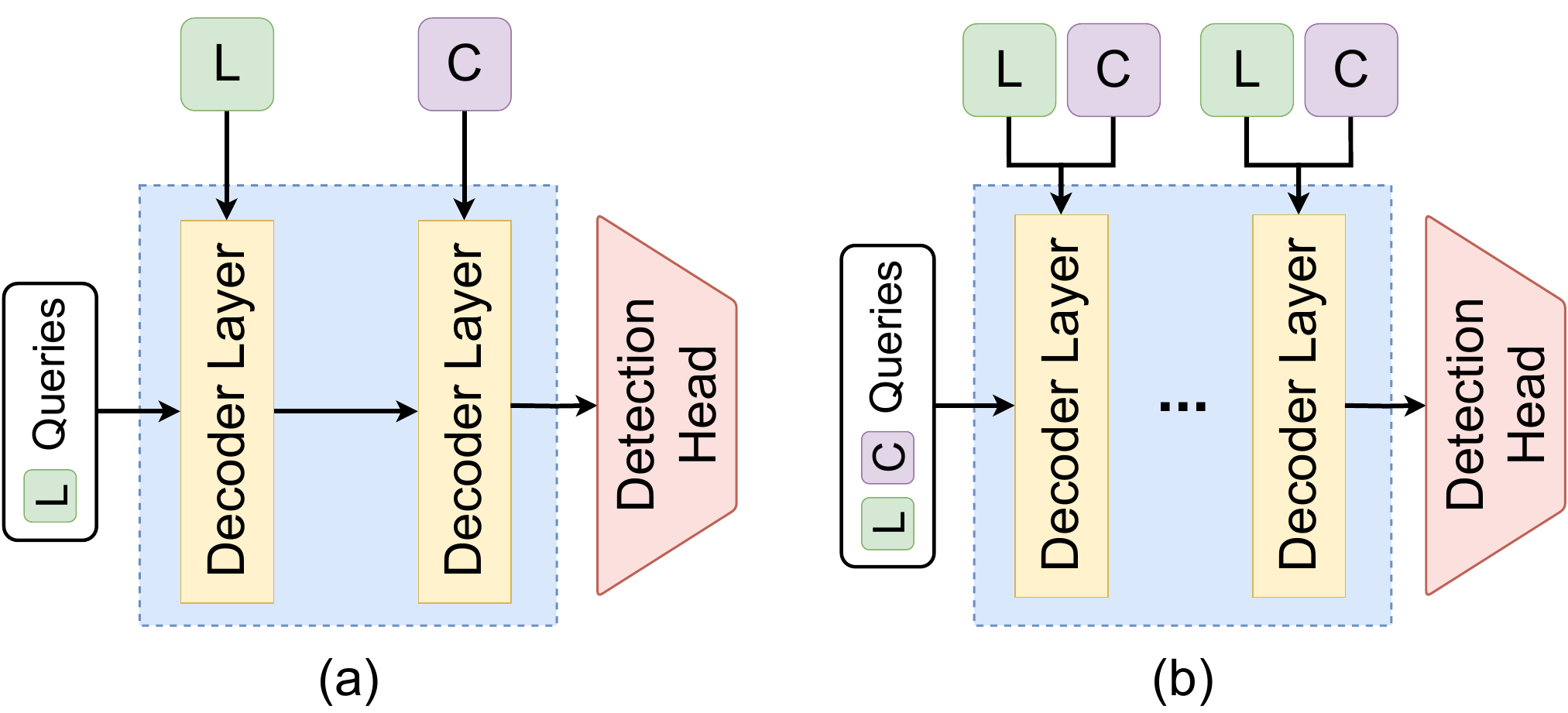}
    \caption{Different approaches to sensor fusion within a transformer decoder. We call (a) sequential fusion, found in TransFusion~\cite{bai_transfusion_2022} and (b) modality-balanced fusion in our proposed method.
    \textbf{\color{L}L} is LiDAR and \textbf{\color{C}C} is camera.}
    \label{fig:fusion_compare}
\end{figure}

Input-agnostic initialization generally requires many queries and multiple passes through the transformer decoder to achieve strong detection performance. 
% Methods with learned object queries need many queries to sufficiently cover the large grid.
% Even then, they still may miss objects if many are located in a small area, because there will not be enough queries close by to cover the objects.
Input-dependent initialization can improve on this  by placing the queries at locations where we expect to find objects after predictions from a first-stage network.
% Experiments in \cite{bai_transfusion_2022} show that strong initialization enables the model to use fewer decoder layers and fewer object queries.
% On top of that, the quality of predictions improves the closer a query is to the corresponding object~\cite{ruppel_transformers_2022-1}.
%TODO set the scope. Why don't we consider BEV fusion?
The concept of input-dependent query initialization in itself is not novel: it was proposed in~\cite{yao_efficient_2021} for the 2D domain and applied in TransFusion~\cite{bai_transfusion_2022} for LiDAR-camera 3D detection.
This state-of-the-art method has also been used as the query initialization approach for other models like DeepInteraction~\cite{yang_deepinteraction_2022}.
We however find two limitations with TransFusion's initialization that lead us to propose a new method.
First, TransFusion takes a sequential approach to sensor fusion, where camera features are only used to refine a set of initial LiDAR-only predictions (\cref{fig:fusion_compare}{a}). 
This limits the benefit of the camera modality.
% Although their query locations are predicted from LiDAR and camera features, the corresponding object query feature vectors are initialized only with LiDAR features and their model does therefore not fully exploit the camera features.
Second, TransFusion uses an elaborate transformer network solely to fuse LiDAR and camera features into a shared bird's eye view (BEV) space from which the initial query locations are predicted. 
This adds significant overhead to the model.
% object query initialization.

To overcome these drawbacks, we propose a novel input-aware initialization approach to initialize object queries with both LiDAR and camera features while introducing only minimal computational overhead.
Our method is combined with a \emph{modality-balanced} transformer decoder (\cref{fig:fusion_compare}{b}), where the object queries can access both sensor modalities in each decoder layer.

Our main contributions are as follows. 
We propose a novel input-dependent object query initialization strategy to initialize the query feature vectors with both LiDAR and camera features in a modality-balanced way, which is also able to work with any other combination of sensor modalities.
We combine the proposed initialization with a modality-balanced transformer decoder to achieve state-of-the-art performance on the nuScenes~\cite{caesar_nuscenes_2020} benchmark.
The proposed initialization outperforms alternative solutions while being more efficient than the query initialization in state-of-the-art method TransFusion~\cite{bai_transfusion_2022}.
\section{Related Work}
% In this section \dots TODO
Our work relates to two main disciplines in 3D object detection: LiDAR-only and LiDAR-camera detection.
In addition, we elaborate on related methods for object query initialization with transformer-based object detection models.

\subsection{LiDAR-based 3D Detection}
\label{sec:lidar-based}
For unimodal models, LiDAR-based detectors top the tables of popular benchmarks like nuScenes~\cite{caesar_nuscenes_2020} and Waymo~\cite{sun_scalability_2020}. 
Most of these detectors quantize the LiDAR point cloud into a regular grid of 3D voxels~\cite{zhou_voxelnet_2018, yan_second_2018}, 2D pillars~\cite{lang_pointpillars_2019} or perspective range views~\cite{fan_rangedet_2021, chai_point_2021, sun_rsn_2021} using convolutional backbones. 
Others operate directly on the unordered, irregular point cloud~\cite{charles_pointnet_2017, qi_pointnet_2017, misra_end--end_2021, yang_3dssd_2020, li_lidar_2021}.
Many detectors adopt anchor boxes in their detection heads~\cite{lang_pointpillars_2019, zhou_voxelnet_2018, deng_voxel_2021}, where objects are predicted as offsets to these anchor boxes. Alternatively, center-based detectors~\cite{yin_center-based_2021, zhou_centerformer_2022} capture objects as points and predict the 3D bounding box from this center point representation.
%TODO add something about 2-stage. heatmap 

After the pioneering work of DETR~\cite{carion_end--end_2020} that applied transformers to 2D detection, transformer models have also found their way to 3D object detection.
Contrary to DETR~\cite{carion_end--end_2020}, where the model directly predicts the bounding box location in global image coordinates, most transformers for 3D detection predict boxes relative to an~\emph{anchor point}. 
% Note that these anchor points are different from anchor boxes discussed above: only the bounding box location is predicted as an offset to the anchor, contrary to both the location and box size.
These anchor locations are either fixed and independent of the current input \cite{chen_futr3d_2023}, or computed using the current point cloud by a sampling method~\cite{misra_end--end_2021, ruppel_transformers_2022} or center heatmap approach~\cite{bai_transfusion_2022, zhou_centerformer_2022, yin_center-based_2021}.

\subsection{LiDAR-Camera 3D Detection}
% LiDAR-camera sensor fusion is widely used in multimodal 3D detectors because of the complementary nature of irregular, sparse 3D point clouds and dense, high-resolution, textured 2D images. 
Early works in LiDAR-camera sensor fusion mainly adopt proposal-level fusion \cite{chen_multi-view_2017, ku_joint_2018}, where proposals are generated in both modalities individually and then shared to the other(s) by projection.
Following PointPainting \cite{vora_pointpainting_2020}, other works similarly apply semantic segmentation networks on images to augment point clouds with richer features \cite{wang_pointaugmenting_2021, huang_epnet_2020, xu_fusionpainting_2021}.
These methods are better able to exploit the multimodal features but are more sensitive to feature alignment issues from suboptimal sensor calibration because of the hard association between points and pixels. 
Finally, there are methods that fuse both modalities into a shared BEV space, either with a direct BEV projection (view transform) of the image pixels \cite{liang_deep_2018,liu_bevfusion_2022, drews_deepfusion_2022, zeng_lift_2022, li_unifying_2022, yoo_3d-cvf_2020}, or by explicitly \emph{lifting} image pixels into 3D space using projected LiDAR depth information \cite{liang_bevfusion_2022, jacobson_center_2022, yin_multimodal_2021, jiao_msmdfusion_2022, yang_deepinteraction_2022}. Within transformer-based models, we see two main approaches for multimodal fusion: 
 those that deploy transformers only as the fusion mechanism for the sensor features~\cite{xu_fusionpainting_2021, yang_ralibev_2022,zeng_lift_2022, kim_3d_2022} 
 and those that use transformers for both sensor fusion and the actual object detection ~\cite{xu_fusionrcnn_2022, yang_deepinteraction_2022, bai_transfusion_2022, chen_futr3d_2023}. The proposed method is based on the latter principle.

\subsection{Object Query Initialization}
% Since our work encompasses object query initialization in transformer-based models, we present an overview of related methods in this section.
Many advances in query initialization originate from 2D object detectors in the image domain. 
In DETR's \cite{carion_end--end_2020} initial implementation, the object queries are a small set of $M=100$ learned embeddings with length $d=256$, which form the object queries~$\{\mathbf{q}_i\}^M_{i=1} \in \mathbb{R}^{d}$.
Deformable DETR~\cite{zhu_deformable_2021} adds positional information to the object queries
% to improve convergence speed and detection performance. 
so that predictions are made as offsets relative to the query positions, rather than globally. 
The object queries~$\{\mathbf{q}_i\}^M_{i=1} \in \mathbb{R}^{d}$ are now accompanied by their 2D locations~$\{\mathbf{c}_i\}^M_{i=1} \in \mathbb{R}^{2}$.
%with $\mathbf{c} \in [0,1]^2$.
In either case, the queries are learned and independent of the current sensor inputs at test time. 
Efficient DETR \cite{yao_efficient_2021} applies input-dependent initialization with a region proposal network (RPN) to transformer-based 2D detection.
% , which again speeds up convergence.

\begin{figure*}[t]
    \centering
    \includegraphics[width=\textwidth]{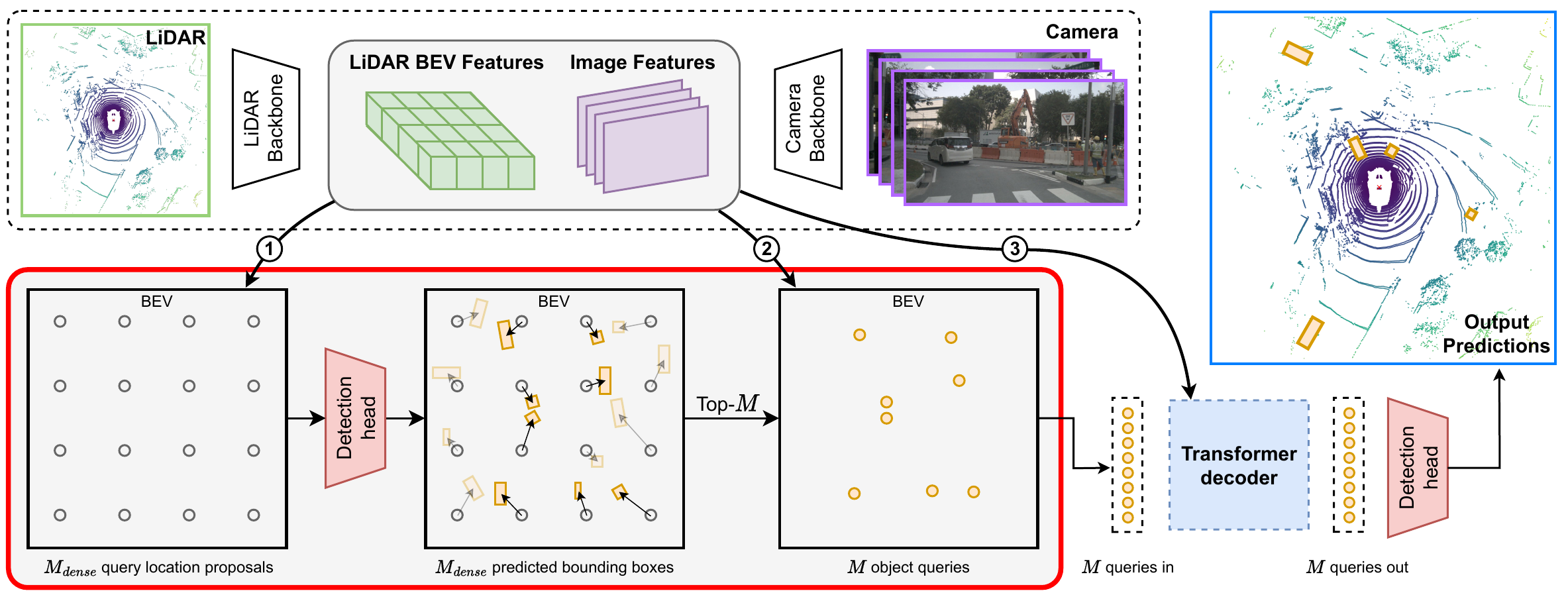}
    \caption{Overview of EfficientQ3M, with the main contribution framed in \textbf{\color{red}red}. We start with a fixed grid $\mathcal{C}$ of $M_{dense}$ query location proposals. We sample LiDAR and camera features at instance level for each proposal \raisebox{.5pt}{\textcircled{\raisebox{-.9pt} {1}}}, and predict a bounding box relative to the grid location. The 3D $xyz$ centers of the top-$M$ bounding boxes with the highest confidence scores are selected as the set of initial object query locations. We re-sample LiDAR and camera features for these $M$ object queries \raisebox{.5pt}{\textcircled{\raisebox{-.9pt} {2}}} and pass them to the modality-balanced decoder, where the queries have access to both sensor modalities in each layer of the decoder \raisebox{.5pt}{\textcircled{\raisebox{-.9pt} {3}}}. A regression and classification head is used to produce the final detections from the object queries at the output of the decoder.}
    \label{fig:model_overview}
\end{figure*}

In 3D object detection, the total area spanned by the scene is very large relative to the size of objects.
% It is desirable to have queries located close to the objects that they are predicting, as this improves the quality of resulting predictions~\cite{ruppel_transformers_2022-1}.
FUTR3D~\cite{chen_futr3d_2023} operates with learned object queries and reference points similar to deformable DETR \cite{zhu_deformable_2021}, but uses 3D locations and increases the number of object queries to $M=900$ to get sufficient coverage of the large space.
Other works~\cite{ruppel_transformers_2022, misra_end--end_2021} use farthest point sampling~\cite{qi_pointnet_2017} on the input point cloud to evenly spread query locations based on the current input. 
%For this, 3D query locations~$\{\mathbf{c}_i\}^M_{i=1} \in \mathbb{R}^{3}$ are sampled from the point cloud, and the corresponding feature vectors~$\{\mathbf{q}_i\}^M_{i=1} \in \mathbb{R}^{d}$ are $d$-dimensional positional embeddings computed from the respective locations.
Finally, there are methods~\cite{bai_transfusion_2022, yang_deepinteraction_2022, zhou_centerformer_2022} most related to ours with input-informed query initialization, where the 2D BEV query locations~$\{\mathbf{c}_i\}^M_{i=1} \in \mathbb{R}^{2}$ are taken as the top-$M$ peaks in a predicted heatmap, and the feature vectors~$\{\mathbf{q}_i\}^M_{i=1} \in \mathbb{R}^{d}$ are initialized with LiDAR features sampled at the locations. 
%Notably, none of these methods initialize object queries $\mathbf{q}$ with camera features.
%Notably, they suffer from increased latency and none of them employ a modality-balanced initialization, which is solved by our proposed method.
Different from them, the proposed method results in 3D initial locations~$\{\mathbf{c}_i\}^M_{i=1}~\in~\mathbb{R}^{3}$, adds minimal overhead while incorporating modality-balanced fusion and initializes the query vectors with both LiDAR and camera features.

\section{Methodology}
In this section, we explain EfficientQ3M, a new multimodal method for input-dependent object query initialization, where object queries are initialized with both LiDAR and camera features sampled at predicted 3D locations. 
% The proposed initialization method is combined with a modality-balanced transformer decoder (\cref{fig:fusion_compare}{b}) following FUTR3D~\cite{chen_futr3d_2023}.
An overview of our proposed model is presented in \cref{fig:model_overview}.

\subsection{Multimodal Input-Dependent Initialization}
\label{sec:dense_init}
% Our approach is motivated as follows. 
In essence, we adopt a lightweight network to predict initial object locations from a large set of $M_{dense}$ proposal object queries, and base the initial object queries for the transformer decoder on the top-$M$ proposals.
We choose $M_{dense}$ to be much larger than $M$ in order to have a sufficient number of proposals to cover the space densely and to not miss any objects. The initialization is lightweight, therefore it can be performed with a large number of $M_{dense}$ queries, whereas the number of queries $M$ in the decoder following the initialization is kept small. 
The proposed method is explained below in more detail.

We start by creating a dense grid of query proposal locations $\mathcal{C} \in \mathbb{R}^{X\times Y\times 1}$, spread out uniformly over the detection range in the $(x,y)$ direction.
Each 2D location in grid $\mathcal{C}$ is assigned the same fixed height to get 3D query locations.
We now have a set of $M_{dense} = X\cdot Y$ initial query proposal locations, which are independent of the current sensor inputs. 
% Given are a LiDAR BEV feature map and a camera feature map from the respective backbones.
For all $M_{dense}$ proposals, we sample sensor features at instance level from the given sensor feature maps after computing the respective view projections (\cref{fig:model_overview} \raisebox{.5pt}{\textcircled{\raisebox{-.9pt} {1}}}) using available intrinsic and extrinsic sensor parameters.
The sensor features are fused, resulting in the corresponding query feature vectors. 
Queries are then made location-aware by adding a positional embedding based on their location using a sine encoding, following DETR~\cite{carion_end--end_2020}.
The query feature vectors now contain the information needed to predict a 3D bounding box relative to their respective locations. 
We predict a classified bounding box from all $M_{dense}$ query proposals, using the same regression head $\Phi_{reg}$ and classification head $\Phi_{cls}$ found later in the transformer decoder layers. 
From all proposal bounding boxes, we select the top-$M$ with the highest confidence scores, and let the queries from which they are predicted be our set of initial object queries.
% , with $M \ll M_{dense}$. 

For each query $\mathbf{q}_i$ out of the top-$M$ proposal queries, the location $\mathbf{c}_i$ is updated with the predicted 3D offset $\Delta \mathbf{x}_i$ from the regression head, i.e. $\mathbf{c}^\prime_i = \mathbf{c}_i + \Delta \mathbf{x}_i$.
We finally generate new query feature vectors from the updated locations, where the query feature vector $\mathbf{q}_i$ is composed by re-sampling features at the new location $\mathbf{c}^\prime_i$ (\cref{fig:model_overview} \raisebox{.5pt}{\textcircled{\raisebox{-.9pt} {2}}}).
This results in the object query feature vectors  $\{\mathbf{q}_i\}^M_{i=1} \in \mathbb{R}^d$ and their 3D locations  $\{ \mathbf{c}^\prime_i\}^M_{i=1} \in \mathbb{R}^3$, which we pass to the decoder. 
In the decoder, the object queries interact through self-attention, and have access to the sensor features from all modalities in cross-attention (\cref{fig:model_overview} \raisebox{.5pt}{\textcircled{\raisebox{-.9pt} {3}}}).

Because we predict a 3D offset $\Delta \mathbf{x}_i$ with $\Phi_{reg}$, we deem it not necessary to initialize a 3D dense grid $\mathcal{C} \in \mathbb{R}^{X\times Y\times Z}$ with multiple different heights as proposals. 
% We conduct experiments where we initialize multimodal query feature vectors from 2D query locations found by TransFusion's~\cite{bai_transfusion_2022} initialization, and find that it performs worse (\cref{sec:abl}).

\paragraph{Modular and Multimodal}
Our initialization method is modular because it can work with any sensor combination, such as camera-only, camera-RADAR, LiDAR-only and LiDAR-camera, similar to the modality-balanced decoder in FUTR3D~\cite{chen_futr3d_2023}. 
The initial grid $\mathcal{C}$ with the proposal locations is identical in each case, but the modalities of features sampled at the locations will differ.
Sensor features are sampled from the corresponding feature map around the projected query location.
If there are multiple sensor modalities available, we concatenate the sampled features from both and fuse them as follows, exemplary for LiDAR-camera fusion:
\begin{equation}
\label{eq:fusion}
\mathcal{S} \mathcal{F}_{\text {fus }}^i=\Phi_{\text {fus }}\left(\mathcal{S} \mathcal{F}_{\text {lid }}^i \oplus \mathcal{S} \mathcal{F}_{\text {cam }}^i\right).
\end{equation}
Here, $\mathcal{SF}_{\text {lid }}^i$ are the sampled LiDAR features, $ \mathcal{S} \mathcal{F}_{\text {cam}}^i$ the camera features, $\Phi_{\text {fus}}$ the fusion multi-layer perceptron (MLP) and $\mathcal{S} \mathcal{F}_{\text {fus}}^i$ is the fused feature vector for query $\mathbf{q}_i$.
When there is only one sensor modality available (e.g. LiDAR-only), $\Phi_{\text {fus}}$ is simply a linear projection. 
% since the dimension of the sampled features already matches that of the object query feature vector $\mathbf{q}_i$.

In contrast to TransFusion~\cite{bai_transfusion_2022}, where an elaborate transformer network is used to create a shared LiDAR-camera heatmap to initialize the object queries, our implementation is lightweight and straightforward.
On top of that, it is flexible and not specific to any sensor suite.
% We show in \cref{sec:results} that the proposed initialization introduces minimal overhead compared to using a learned distribution for the queries.

\paragraph{Model Details}
% We combine the proposed object query initialization method with a modality-balanced transformer decoder like in FUTR3D~\cite{chen_futr3d_2023}.
The implementation of the backbones, transformer decoder, and final detection head -- i.e. all components outside of the red frame in \cref{fig:model_overview} -- follows FUTR3D~\cite{chen_futr3d_2023}.
Our initialization method may be applied to other decoder designs with any combination of sensor modalities, as long as there exist transformations from global 3D coordinates to the respective sensor feature map coordinates.

\subsection{Losses}
\label{sec:loss}
We supervise the model in three locations, starting with the object query initialization.
As explained in \cref{sec:dense_init}, we predict a large set of $M_{dense}$ bounding boxes and initialize our object queries from the top-$M$ with the highest confidence score.
To supervise the $M_{dense}$ bounding boxes, we perform bipartite matching between the ground truth objects and all $M_{dense}$ predicted boxes to get a set of one-to-one matches.
The Hungarian algorithm~\cite{kuhn_hungarian_1955} is used to produce these matches.
The associated matching cost is a weighted sum of classification and regression costs:

\begin{equation}
\label{eq:matching_cost}
    C_{match}=\lambda_1 L_{c l s}(\hat{p},p)+\lambda_2 L_{r e g}(\hat{b}, b)
\end{equation}

where $(\hat{p},\hat{b})$ are the predicted class confidence scores and bounding box parameters and $(p,b)$ the corresponding supervisory signals, $L_{cls}$ is the focal loss~\cite{lin_focal_2017} and $L_{reg}$ the L1 regression loss, and $\lambda_1, \lambda_2$ are the corresponding weights.
% with $\lambda_1 = 2$ and $\lambda_2 = 0.25$. 
For the set of matched predictions, we again compute the classification loss (focal loss) and regression loss (L1 loss).

Additionally, we supervise all $M_{dense}$ predictions with a dense heatmap to improve convergence, because the number of matched predictions is much smaller than the number of proposals $M_{dense}$.
For this, we take the class confidence scores from all predictions as a class-specific dense heatmap $\hat{\mathcal{S}} \in \mathbb{R}^{X\times Y \times K}$, which is supervised by a ground truth heatmap $\mathcal{S} \in \mathbb{R}^{X\times Y \times K}$ with the penalty reduced focal loss.  
Here, $X\times Y$ defines the spatial dimension of the heatmap, which matches our dense grid of query locations $\mathcal{C} \in \mathbb{R}^{X\times Y}$ and $K$ is the number of classes. 
We take the $K$ confidence scores for each prediction from grid $\mathcal{C}$ to obtain heatmap $\hat{\mathcal{S}}$.
The ground truth heatmap is computed following CenterPoint~\cite{yin_center-based_2021}.
We find that without this dense loss term, our method does not converge. 

Next, the predicted bounding boxes at the output of each transformer decoder layer are supervised as in FUTR3D~\cite{chen_futr3d_2023}.
Finally, because sparse supervision can hinder learning in transformer-based models~\cite{zong_detrs_2023}, we follow FUTR3D and implement an auxiliary detection head parallel to the transformer decoder for improved supervision of the LiDAR backbone. 
This CenterPoint~\cite{yin_center-based_2021} head is only used to help the LiDAR backbone learn better features during training, and is removed at test time. 

\begin{table*}[t]
\centering
\small
\caption{Comparison to the state of the art on the nuScenes \emph{validation} and \emph{test} set. The best scores for each sensor modality are \textbf{bold}.}
\label{tab:nuscenes_test}
\begin{adjustbox}{max width=\textwidth}%
\begin{tabular}{l|c|cc|ccll}
\toprule
Method          & Modality & \multicolumn{2}{c|}{Backbone} & \multicolumn{2}{c}{\emph{validation}} & \multicolumn{2}{c}{\emph{test}} \\
                &          & Camera       & LiDAR          & mAP $\uparrow$    & NDS $\uparrow$    & mAP $\uparrow$ & NDS $\uparrow$ \\ 
\hline 
\noalign{\vskip 0.3ex} 
CenterPoint~\cite{yin_center-based_2021}     & L        & -            & VoxelNet       & 59.6              & 66.8              & 60.3           & 67.3           \\
TransFusion-L~\cite{bai_transfusion_2022}   & L        & -            & VoxelNet       & 65.1           & \textbf{69.9}              & 65.5           & 70.2           \\
% \rowcolor{gray1}
FUTR3D~\cite{chen_futr3d_2023}          & L        & -            & VoxelNet       & 63.7              & 69.0              & 65.3   & 69.9           \\
% \rowcolor{gray2}
EfficientQ3M   (ours)          & L        & -            & VoxelNet       & \textbf{65.3}              & 69.6    &      \textbf{66.1}          &    \textbf{70.3}            \\ 
% [0.5ex] 
\hline \noalign{\vskip 0.3ex} 
% MVP~\cite{yin_multimodal_2021}            & L+C      & DLA34        & VoxelNet       & 67.1              & 70.8              & 66.4           & 70.5           \\
PointAugmenting~\cite{wang_pointaugmenting_2021} & L+C      & DLA34        & VoxelNet       & -                 & -                 & 66.8           & 71.0           \\
TransFusion~\cite{bai_transfusion_2022} & L+C      & DLA34          & VoxelNet       & 67.3              & 70.9              & 68.9           & 71.6           \\
DeepInteraction~\cite{yang_deepinteraction_2022}          & L+C      & R50       & VoxelNet       & 69.9            & 72.6              & \textbf{70.8}           & \textbf{73.4}           \\
% \rowcolor{gray1}
FUTR3D~\cite{chen_futr3d_2023}          & L+C      & VoVNet       & VoxelNet       & 70.3              & 73.1     & 69.4           & 72.1           \\
% \rowcolor{gray2}
EfficientQ3M (ours)            & L+C      & VoVNet       & VoxelNet       & \textbf{71.2}              & \textbf{73.5}              &     70.5           &    72.6            \\ 
\bottomrule
\end{tabular}
\end{adjustbox}
\end{table*}

\section{Experiments}
In this section, we introduce the chosen dataset and implementation details before presenting the main results and ablation studies.
% In this section we elaborate on the chosen dataset and implementation details before presenting the main results.

\subsection{Dataset and Metrics}
We use the large-scale autonomous driving dataset nuScenes~\cite{caesar_nuscenes_2020} to evaluate our model.
The dataset is a collection of $1000$ \emph{scenes}, each $20\thinspace$seconds long and annotated at $2\thinspace$Hz. 
Objects are annotated with 3D bounding boxes and a class label out of $K=10$ object classes. 
Performance is measured with the popular mean average precision (mAP) metric and the custom nuScenes detection score (NDS). 

\subsection{Implementation Details}
Our implementation is written using PyTorch~\cite{paszke_automatic_2017} in the open-source MMDetection3D~\cite{mmdetection3d_contributors_openmmlabs_2020} framework.
The code is built on top of the public release of FUTR3D~\cite{chen_futr3d_2023}, and we follow their model settings for the hyperparameters, unless stated otherwise.
The most important details are listed below. 

\subsubsection{Model Settings} 
% The voxel size of the LiDAR backbone is $(0.075\thinspace\text{m}, 0.075\thinspace\text{m}, 0.2\thinspace\text{m})$.
The LiDAR backbone is a VoxelNet~\cite{yan_second_2018, zhou_voxelnet_2018} with a voxel size of $(0.075\thinspace\text{m}, 0.075\thinspace\text{m}, 0.2\thinspace\text{m})$.
The image backbone is a VoVNET~\cite{lee_energy_2019} and is pre-trained on nuScenes~\cite{caesar_nuscenes_2020} using the camera-only version of FUTR3D. 
% The image backbone is pretrained on nuScenes~\cite{caesar_nuscenes_2020} using the camera-only version of FUTR3D. 
%The detection range is $[-54\thinspace\text{m},54\thinspace\text{m}]$ for the $X,Y$ grid and $[-5\thinspace\text{m}, 3\thinspace\text{m}]$ for the $Z$ axis.
% The number of levels in the feature pyramid networks of the LiDAR and camera backbones is $m=4$.
% The channel dimension of the sensor feature maps and the object queries is $d=256$.
% The number of sampling offsets in the deformable attention is $V=4$.
% Like in nuScenes, the number of camera images is $N=6$ and the number of object classes is $K=10$.
The number of object query proposals is $M_{dense}=3600$, spread out uniformly over the $(x,y)$ detection range.
We report results for both $M=200$ and $900$ object queries.%, with $M$ the number of object queries that are input to the decoder.

\subsubsection{Training Strategy}
We adopt a common augmentation pipeline for the LiDAR data, where we use random rotation with $r \in [\pi/4, \pi/4]$, random scaling with $s \in [0.9,1.1]$, random $xyz$ translation with a standard deviation of $0.5$, and random horizontal and vertical flipping.
We use CBGS~\cite{zhu_class-balanced_2019} class-balanced sampling to improve the class balance in nuScenes.
Finally, we use ground truth copy-paste augmentation~\cite{yan_second_2018}, and disable it for the final epochs to match the real data distribution again~\cite{wang_pointaugmenting_2021}.
We do not adopt any augmentation at test time.

We first train the LiDAR branch of our model, using a pre-trained LiDAR backbone. 
The schedule is set to 6 epochs, with GT copy-paste augmentation disabled in the final 3 epochs. 
From there, the pre-trained image backbone is added and the LiDAR-camera model is trained for another 6 epochs with both backbones frozen. 
Such a sequential approach has shown to yield better performance than joint training from the start, because it allows for better augmentation in the LiDAR-only stage of training~\cite{chen_futr3d_2023, bai_transfusion_2022}.
We use an initial learning rate of $1.0\times 10^{-4}$ for both LiDAR-only and LiDAR-camera training, with a cyclic learning rate policy \cite{smith_cyclical_2017}.

\begin{figure}[t]
    \centering
    \includegraphics[width=1\linewidth]{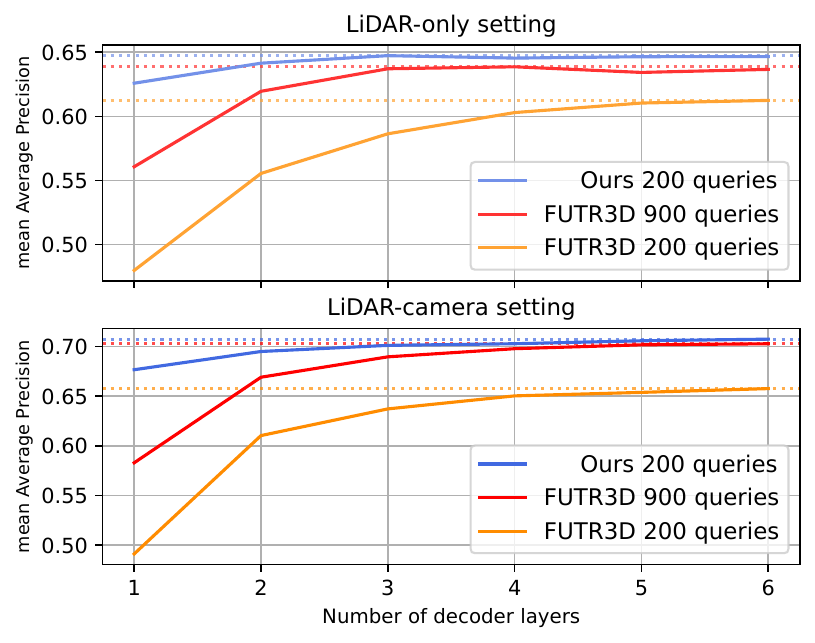}
\caption{Detection performance vs. the number of decoder layers on the nuScenes \emph{val} set in the LiDAR-only and LiDAR-camera setting, compared to FUTR3D~\cite{chen_futr3d_2023}. The proposed method needs fewer layers and fewer queries to outperform FUTR3D.
%We show LiDAR-only detection on top and LiDAR-camera detection below. 
% We compare our proposed method with $200$ queries to the FUTR3D~\cite{chen_futr3d_2023} baseline with both $900$ and $200$ queries.
}
    \label{fig:decoder_layers}
\end{figure}
% The reason for the score discrepancy is unknown.

\begin{figure*}[t]
    \centering
    \includegraphics[width=1.0\textwidth]{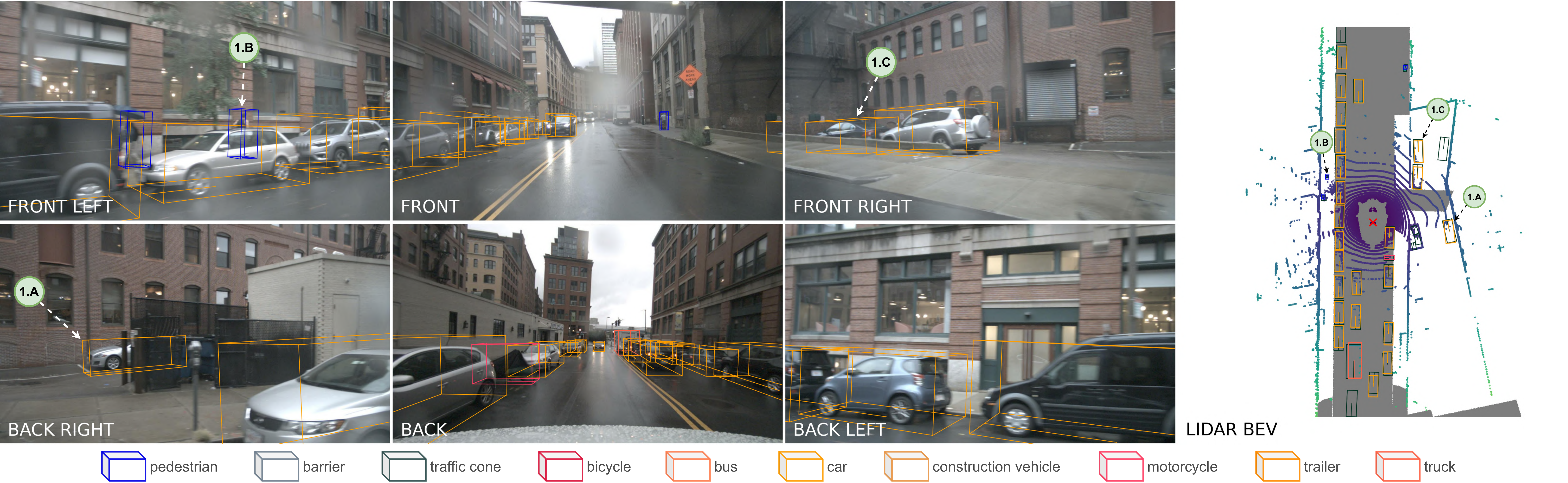}
    \caption{Example of output predictions with the proposed model on the nuScenes \emph{val} set. The LiDAR BEV shows ground truth objects in dark green.}
    \label{fig:viz_good}
\end{figure*}

\subsection{Main Results}
\label{sec:results}

The detection performance of EfficientQ3M is evaluated on both the nuScenes \textit{validation} and \textit{test} set and the results are presented in \cref{tab:nuscenes_test}.
On the nuScenes \textit{val} set, the proposed method outperforms the state of the art in transformer-based object detection for both the LiDAR-only, as well as the LiDAR-camera fusion setting. This also applies for the \textit{test} set with LiDAR-only detection. With LiDAR-camera fusion on the \textit{test} set, EfficientQ3M outperforms its baseline FUTR3D \cite{chen_futr3d_2023} as well as TransFusion \cite{bai_transfusion_2022}. Note that, since DeepInteraction \cite{yang_deepinteraction_2022} utilizes TransFusion's initialization strategy, it may further benefit by incorporating the proposed modality-balanced method for intialization.
% However, we find that these results do not translate to the \emph{test} set.
% Concerning the test set, we differentiate between two different scores obtained by the FUTR3D baseline: those stated in their paper are listed first, while those obtainable with their publicly available model weights are marked with an asterisk. 
% This discrepancy is likely caused by a difference in training strategy.
% FUTR3D's top results are obtained after training on the nuScenes training \emph{and} validation set, while those marked with an asterisk are not.
% Our proposed method is trained only on the training set, and is able to outperform the corresponding* FUTR3D scores. 

In \cref{fig:decoder_layers}, we compare our model to FUTR3D with a varying number of decoder layers. 
We find that the proposed input-dependent object query initialization not only produces superior detection scores, but also enables the use of fewer object queries and decoder layers.
% The benefit of input-dependent query initialization is clearly visible here. 
Where FUTR3D needs multiple passes through the decoder to achieve high performance, our input-dependent initialization already shows good performance for a single-layer model.
Specifically, the proposed method with $200$ object queries outperforms the baseline with $900$ queries for any number of decoder layers.
We also find that the input-agnostic baseline degrades strongly when decreasing the number of object queries.

\begin{table}[t]
    % \small
    \centering
    \caption{Comparison of different query initialization methods paired with the modality-balanced decoder on the nuScenes \emph{val} set. 
    }
    \begin{tabular}{l|ll|lll}
        \toprule
        Init. Method & Mod. & \#Q  & mAP $\uparrow$ & Lat. (ms) $\downarrow$  & \#P   \\ \midrule
        Input-agnostic   & L&900 &63.7 & 208.6 & 7.29      \\
        w/ TransFusion & L& 200& 64.5 \textsubscript{\color{black}(+0.8)} & 212.7 \textsubscript{\color{red}(+2.0\%)} & 7.89  \\
        w/ ours & L& 200& 64.7 \textsubscript{\color{blue}(+1.0)} & 209.5 \textsubscript{\color{black}(+0.4\%)} & 7.51  \\ 
        [0.3ex] \hdashline \noalign{\vskip 0.3ex}
        Input-agnostic   & L+C& 900 & 70.3& 645.4 & 9.32    \\
        w/ TransFusion & L+C& 200&70.3 \textsubscript{\color{black}(+0.0)}& \cellcolor{black!6}751.1 \textsubscript{\color{red}(+16.4\%)} & 11.8  \\
        w/ ours & L+C& 200 & 70.8 \textsubscript{\color{blue}(+0.5)}& \cellcolor{black!6}652.3 \textsubscript{\color{black}(+1.1\%)} & 9.80 \\
        \bottomrule
    \end{tabular} 
\label{tab:init_compare}
\end{table}

% The performance improvement with a single decoder layer model is larger for LiDAR-camera fusion (+9.4~mAP) than it is for LiDAR-only detection (+6.5~mAP). 
% We hypothesise that this is due to the height of the query location coming into effect.
% For the projection from query location to image coordinates, the $z$ location is also of importance, which is not the case for sampling from a LiDAR BEV feature map.
% This additional dimension makes it less likely for the learnt query locations in FUTR3D to sample relevant image features in the first decoder layer. 
% Our input-dependent initialization predicts an $xyz$ location for each object query to enable improved sampling of image features, which explains why the single layer performance is still strong even with the additional dimension in the query location. 

In \cref{tab:init_compare} we compare the proposed method to the initialization of the closest related work: TransFusion~\cite{bai_transfusion_2022}, with $M=200$ queries. 
We include FUTR3D's~\cite{chen_futr3d_2023} input-agnostic approach for reference, with $M=900$ queries for a fair comparison.
Only the object query initialization method is varied, all other model settings, such as the backbones, transformer decoder and detection head are equal.   
We find that both initialization strategies achieve similar improvements on the baseline in the LiDAR-only setting, but that the proposed method is superior for LiDAR-camera fusion.
We contribute this to the benefit of our modality-balanced design and to predicting initial 3D query locations, which allow for more accurate sampling of image features, compared to the 2D approach in TransFusion. 

Additionally, we find that the proposed method is lightweight and efficient compared to TransFusion's initialization. 
Especially for LiDAR-camera fusion, our method remains efficient with only $+6.9\thinspace$ms of added latency while TransFusion's introduces $+105.7\thinspace$ms of overhead compared to a model with input-agnostic queries (i.e. a $15\times$ difference in the initialization stage, highlighted in gray).
This is explained by the elaborate transformer network used in their query initialization, the size of which also shows when looking at the number of model parameters \#P (measured in millions, excluding the backbones).   

\cref{fig:viz_good} shows a sample of detections made with the proposed method.
We highlight three successful detections in a difficult setting, i.e. with rain and strong occlusions.

\subsection{Ablation}

\subsubsection{Query Initialization}
We test if it suffices to initialize the query feature vectors with positional embeddings based on their predicted location (Refs.), or if we instead need to initialize them with sensor features sampled at their location (Refs. + Feats.), as proposed.
We compare both approaches with the input-agnostic baseline for $M=200$ object queries, with all methods trained on a reduced schedule, see \cref{tab:lidar_abl}.

\begin{table}[t]
\centering
% \small
\caption{Ablation on object query initialization  components on the nuScenes \emph{val} set, in the LiDAR-only setting with $200$ queries. }
\begin{tabular}{l|c|cc}
\toprule
Query Initialization Strategy & \#L. & mAP $\uparrow$ & NDS  $\uparrow$ \\ \midrule
Input-agnostic & 6            & 60.9 $\pm$ 0.2           & 67.1 $\pm$ 0.1            \\
Input-dependent Refs.  & 6            & 64.3 $\pm$ 0.1           & \textbf{69.1} $\pm$ 0.1           \\
Input-dependent Refs. + Feats.    & 6            & \textbf{64.3} $\pm$ 0.1           & 69.0 $\pm$ 0.0            \\ 
\hdashline \noalign{\vskip 0.3ex}
Input-agnostic  & 1            & 48.2 $\pm$ 0.2           & 56.8 $\pm$ 0.2            \\
Input-dependent Refs.  & 1            & 61.8 $\pm$ 0.2           & 67.0 $\pm$ 0.2           \\
Input-dependent Refs. + Feats. & 1            & \textbf{62.2} $\pm$ 0.1          & \textbf{67.3} $\pm$ 0.1            \\ \bottomrule
\end{tabular}
\label{tab:lidar_abl}
\end{table}

We find that, for 6 decoder layers, both versions of input-dependent initialization perform similarly: initializing the query feature vectors with sensor features does not bring obvious benefits.
When we decrease the number of decoder layers to 1, however, the benefit of initializing the query vectors with sensor features is visible: it results in a small performance increase compared to using the positional embedding.
Additionally, we see the improvement on the input-agnostic baseline increase.
The baseline performance drops heavily because it needs multiple layers to iteratively update the query location if it is not located close to an object by chance at initialization. 
% Notably, our approach only shows a relatively small drop of $2.1\thinspace$mAP. 

\subsubsection{Number of Queries}
FUTR3D~\cite{chen_futr3d_2023} uses $M=900$ to get sufficient coverage of the large 3D space. 
With the proposed input-dependent query initialization, we are able to use fewer queries and still have them located close to the objects in the scene. 
We compare our method with FUTR3D for $200$ and $900$ queries, where we have fine-tuned FUTR3D's pretrained model to learn a new distribution with $M=200$. 

In \cref{tab:num_query_abl}, we find that the proposed method can still achieve strong performance even with many fewer queries, thanks to the input-dependent initialization.
Specifically, our method with $200$ queries outperforms FUTR3D with $900$ queries.
We do still see a marginal benefit of using $900$ queries with the proposed initialization method, even though $200$ queries are enough to cover the maximum number of objects in nuScenes.
We hypothesize that this is caused by the larger self-attention operation in the decoder which allows for querying more information, and by the additional queries compensating for imperfect initialization.

\begin{table}[t]
\centering
% \small
\setlength{\tabcolsep}{8pt}
\caption{Comparing detection performance (mAP $\uparrow$) on the nuScenes \emph{val} set for both sensor setups with $200$ and $900$ object queries.}
\label{tab:num_query_abl}
\begin{tabular}{l|l|cc}
\toprule
   & Mod. & \# 200         & \# 900  \\ \midrule
FUTR3D  &L& 61.3 \textsubscript{\color{red}(-2.4)} & 63.7 \\
FUTR3D &L+C& 65.8 \textsubscript{\color{red}(-4.5)} & 70.3 \\ 
[0.3ex] \hdashline \noalign{\vskip 0.3ex}
EfficientQ3M (ours)     &L& 64.7 \textsubscript{(-0.6)}        & 65.3    \\
EfficientQ3M (ours)     &L+C& 70.8 \textsubscript{(-0.4)}          & 71.2    \\ \bottomrule
\end{tabular}
\end{table}

\section{Conclusion}
\label{sec:discussion}

We introduced EfficientQ3M, a novel and efficient approach for initializing object queries in transformer-based 3D object detection models from any sensor modality.
Existing methods strongly rely on a LiDAR-only first stage, which limits the benefit from the camera domain. %limited the model's recall to that of the LiDAR-only stage.
EfficientQ3M overcomes this limitation by initializing object queries with features from any combination of sensor modalities. Compared to an input-agnostic method, it achieves better performance with fewer queries and decoder layers.
The proposed method, when combined with a modality-balanced transformer decoder, outperforms the state of the art for query-based LiDAR object detection in the nuScenes benchmark.
Additionally, EfficientQ3M demonstrates significant efficiency gains compared to related work for query initialization.
Furthermore, the modularity of the proposed method allows its application to different transformer decoders as well.
% One point of discussion is the general benefit of input-dependent initialization. 
% It is expected that such initialization enables a reduction of the number of object queries and the number of transformer decoder layers while achieving similar performance to a model with input-agnostic object queries (i.e. improving efficiency). 
% We however find that there is also a performance improvement, even though the input-agnostic queries in FUTR3D~\cite{chen_futr3d_2023} should be able to detect the same objects.
% We hypothesize that this advantage is caused by densely populated areas. 
% The learned queries in FUTR3D cover $13\thinspace$m\textsuperscript{2} on average. 
% If there are many objects in a small area, there will not be enough queries around to detect all of them.
% The proposed method is less sensitive to this problem.
For future work, it would be interesting to extend the scope of the experiments to obtain results for more sensor setups, such as RADAR-camera and camera-only.
%Additionally, the proposed initialization can be applied to different decoder designs like DeepInteraction~\cite{yang_deepinteraction_2022}.
% Future work, use BEVFusion features?



% \section*{ACKNOWLEDGMENT}

%%%%%%%%%%%%%%%%%%%%%%%%%%%%%%%%%%%%%%%%%%%%%%%%%%%%%%%%%%%%%%%%%%%%%%%%%%%%%%%%

\bibliographystyle{bibtex/IEEEtran}
\IEEEtriggeratref{44}
\bibliography{bibtex/IEEEabrv, refs}

\end{document}